\documentclass[sigconf]{acmart}

\usepackage{graphicx}
\usepackage{booktabs}
\usepackage{multirow}
\usepackage{siunitx}
\usepackage{subcaption}
\usepackage{amsmath}

\usepackage{amssymb}
\usepackage{float}
\usepackage{enumitem}
\usepackage{xcolor}

\newcommand{\fone}{F1}
\newcommand{\iou}{IoU}

\setcopyright{none}
\renewcommand\footnotetextcopyrightpermission[1]{}
\acmDOI{}
\acmISBN{}
\acmConference{}
\acmYear{}
\acmPrice{}

\begin{document}

\title{A Boundary-Metric Evaluation Protocol for\\Whiteboard Stroke Segmentation Under Extreme Imbalance}

\author{Nicholas Korcynski}
\affiliation{
  \institution{Rowan University}
  \country{USA}
}
\email{korcyn28@students.rowan.edu}
\email{thelostamount@gmail.com}

\begin{abstract}
The binary segmentation of whiteboard strokes is hindered by extreme
class imbalance, caused by stroke pixels that constitute only $1.79\%$ of the image
on average, and in addition, the thin-stroke subset averages
$1.14\% \pm 0.41\%$ in the foreground.
Standard region metrics (\fone{}, \iou{}) can mask thin-stroke
failures because the vast majority of the background dominates the score. In contrast, adding boundary-aware metrics and a thin-subset equity analysis
changes how loss functions rank and exposes hidden trade-offs.
We contribute an evaluation protocol that jointly examines region
metrics, boundary metrics (BF1, B-IoU), a core/thin-subset equity
analysis, and per-image robustness statistics (median, IQR,
worst-case) under seeded, multi-run training with non-parametric
significance testing.
Five losses---cross-entropy, focal, Dice, Dice+focal, and
Tversky---are trained three times each on a DeepLabV3-MobileNetV3
model and evaluated on 12 held-out images split into \emph{core}
and \emph{thin} subsets.
Overlap-based losses improve \fone{} by more than 20 points over
cross-entropy ($0.663$ vs.\ $0.438$, $p < 0.001$). In addition, the boundary metrics
confirm that the gain extends to the precision of the contour.
Adaptive thresholding and Sauvola binarization at native resolution
achieve a higher mean \fone{} ($0.787$ for Sauvola) but with
substantially worse worst-case performance (\fone{} $= 0.452$
vs.\ $0.565$ for Tversky), exposing a
consistency--accuracy trade-off: classical baselines lead on mean
\fone{} while the learned model delivers higher worst-case reliability.
Doubling training resolution further increases \fone{} by 12.7 points.
\end{abstract}

\maketitle

\section{Introduction}
\label{sec:intro}

Whiteboard digitization is a common step in hybrid learning and
collaborative work environments: a photograph of a whiteboard is
processed to extract clean stroke masks that can be imported into
note-taking applications such as Microsoft OneNote.
Although the task is conceptually straightforward, requiring only separation of foreground ink from a uniform background, it is surprisingly challenging in practice. Pen and marker strokes are thin, occupy only a small fraction of the image
area, and vary in color, thickness, and contrast depending on the type of marker, lighting conditions, and camera angle.

\paragraph{Extreme class imbalance.}
Across the 34 original images in the dataset used in this study, the stroke
pixels represent a mean of only $1.79\%$ of the image (range $0.52$--$4.94\%$).
A subset of five images containing especially thin strokes has a mean
stroke fraction of just $1.14\% \pm 0.41\%$ and a mean stroke width of
$11.3 \pm 0.9$\,px, compared to $2.41\% \pm 1.11\%$ and
$21.6 \pm 18.1$\,px for the seven images with core-stroke.
Under such imbalance, a trivial classifier that predicts every pixel as
background achieves $>98\%$ pixel accuracy, making standard cross-entropy
loss an unreliable training signal.

\paragraph{Thin-structure failures.}
Thin strokes are the category that is the most susceptible to failure.
Downsampling during training erodes fine details, and standard losses
assign equal weight to every correctly classified pixel, a regime in
which the overwhelming background majority dominates the gradient.
As a result, models trained with cross-entropy learn to under-predict
thin strokes, achieving high pixel accuracy while missing precisely the
content that matters to the end user. 

\paragraph{Standard vs.\ boundary metrics.}
Region-based metrics such as \fone{} and \iou{} can hide boundary
quality issues: a prediction may achieve moderate overlap with the ground
truth, yet produce ragged or dilated contours.
Boundary-aware metrics, such as boundary F1 (BF1)~\cite{csurka2013} and Boundary IoU (B-IoU)~\cite{cheng2021boundary}, restrict the evaluation to a narrow band surrounding object contours and are therefore more appropriate for quantifying the fidelity of thin structures.
\paragraph{Contributions.}
The primary contribution is not that overlap-based losses outperform
cross-entropy, which is well-established, but rather a rigorous,
reproducible evaluation protocol for thin-structure segmentation under
extreme foreground sparsity. Specifically:
\begin{enumerate}[leftmargin=*,itemsep=1.0ex]
  \item A \textbf{thin subset, boundary-aware evaluation protocol}
        that jointly reports region metrics (\fone{}, \iou{}),
        boundary metrics (BF1, B-IoU), and a core/thin equity analysis,
        revealing performance gaps invisible to region metrics alone.
  \item \textbf{multi-run seed training with non-parametric
        significance testing} (three seeds $\times$ Wilcoxon signed-rank
        with Bonferroni correction), paired with per-image robustness
        statistics (median, IQR, worst-case), which quantify both
        statistical reliability and practical stability.
  \item A \textbf{core vs. thin equity measure} that exposes how
        loss functions differ in their treatment of fine versus thick
        strokes---a diagnostic absent from prior loss-comparison studies.
  \item A \textbf{consistency--accuracy trade-off analysis} between
        learned models and classical baselines (adaptive and Sauvola
        thresholding), demonstrating that the baselines' higher mean
        \fone{} comes at the cost of substantially worse worst-case
        performance.
  \item A \textbf{fully reproducible experimental pipeline} with
        publicly available code and evaluation scripts.
\end{enumerate}

\section{Related Work}
\label{sec:related}

\subsection{Document and Whiteboard Segmentation}
Whiteboard content extraction has been studied as a special case of
document binarization.
Early systems relied on color-space thresholding and connected-component
analysis~\cite{he2005whiteboard,yin2007whiteboard}.
More recently, encoder--decoder architectures such as
U-Net~\cite{ronneberger2015unet} and DeepLabV3~\cite{chen2017rethinking}
have been applied to document segmentation tasks, including
text-line extraction, table detection, and handwriting
isolation~\cite{he2017mask,long2015fully}.
However, most deep-learning studies target printed documents or scene
text rather than the sparse, thin strokes typical of whiteboard
handwriting.

\subsection{Thin-Structure Segmentation}
Segmentation of thin or highly elongated structures, such as blood vessels, road networks, neural fibers, and surface cracks, exhibits the same fundamental challenges as whiteboard stroke delineation, namely, pronounced class imbalance and pronounced sensitivity to boundary localization accuracy.
Topology-aware losses such as clDice~\cite{shit2021cldice} explicitly
enforce connectivity by penalizing topological breaks in tubular
predictions; Skeletal-based evaluation
metrics~\cite{zheng2021rethinking} complement them on the evaluation
side.
The curvilinear-structure literature consistently reports that standard
cross-entropy under-segments thin branches and that overlap-based losses
such as Dice partially mitigate the problem~\cite{milletari2016vnet}.
Recent crack and defect-detection studies~\cite{liu2023crackseg,
pantoja2022topo} confirm this pattern in civil-infrastructure imagery,
where foreground ratios are similarly low ($<$5\%).
We do not include clDice in our ablation because our binary masks lack
per-stroke instance or skeleton annotations; however, it is a natural
extension (Section~\ref{sec:conclusion}).

\subsection{Loss Functions for Class Imbalance}
Several loss functions have been proposed to improve segmentation under
class imbalance.
Dice loss~\cite{milletari2016vnet} directly optimizes the Dice
coefficient, making it insensitive to the number of true-negative pixels.
Generalized Dice~\cite{sudre2017generalised} extends Dice to multi-class settings by weighting each class inversely to its volume and reduces to standard Dice in the binary case.
Focal loss~\cite{lin2017focal} reduces the weight of well-classified pixels so
that the gradient is dominated by hard examples near decision boundaries.
The Tversky loss~\cite{salehi2017tversky} generalizes Dice by introducing
separate weights for false positives and false negatives, allowing a
bias towards recall.
Combinations of Dice and focal loss have been shown to outperform either
component alone in medical-image segmentation~\cite{yeung2022unified}.
Despite this body of work, systematic comparisons under the specific
imbalance ratios found in whiteboard segmentation ($\sim$2\% foreground)
are lacking, and few studies pair loss ablations with boundary-aware
metrics, per-image robustness profiling, or thin-subset equity
analysis, which is the gaps this work addresses.

\subsection{Modern Architectures}
Transformer-based segmentation models such as
SegFormer~\cite{xie2021segformer} and
Mask2Former~\cite{cheng2022mask2former} now achieve state-of-the-art
results on standard benchmarks.  Their larger receptive fields may better
capture global context, but they also require substantially more data
and computation.  We deliberately choose a lightweight CNN backbone
(MobileNetV3) to isolate the effect of the loss function from that of
the architecture, and evaluating modern backbones under the same protocol is
left for future work.

\subsection{Boundary-Aware Evaluation}
Region-based metrics such as \iou{} and Dice can mask boundary errors
when the interior of the object is large.
Boundary F1 (BF1)~\cite{csurka2013} restricts precision and recall
computation to a morphological band around the contour of ground-truth.
The boundary IoU~\cite{cheng2021boundary} similarly restricts the IoU
computation to a 2\% diagonal band, providing a metric that is
particularly sensitive to contour placement.
These metrics have seen increasing adoption in instance and panoptic
segmentation benchmarks but remain uncommon in document analysis.

\subsection{Boundary-Aware and Topology-Aware Losses}
Several losses directly optimize the boundary or topological quality.
InverseForm~\cite{borse2021inverseform} learns a differentiable
approximation of the IoU boundary and has shown gains in Cityscapes and
NYUDv2.
clDice~\cite{shit2021cldice} penalizes topological breaks in tubular
structures by computing Dice in skeletonized predictions.
Boundary loss~\cite{kervadec2019boundary} reformulates
contour-distance matching as a differentiable integral over the
softmax output.
These approaches are complementary to the region-based losses compared
in this study; our evaluation protocol can accommodate them directly,
and we identify them as a priority for future ablation
(Section~\ref{sec:conclusion}).

\section{Method}
\label{sec:method}

\subsection{Dataset}
\label{sec:dataset}

The data set consists of 34 real whiteboard photographs captured with
smartphone cameras in classroom and office settings.
The images have two native resolutions: $3712 \times 2784$ (30~images) and
$3968 \times 2232$ (4~images).
The binary masks of ground-truth were manually annotated with
GIMP\footnote{\url{https://www.gimp.org/}} by a single annotator;
each pixel is labeled as \emph{stroke} (255) or \emph{background} (0).

The stroke coverage ranges from $0.52\%$ to $4.94\%$ of the image area
(mean $1.79\%$, median $1.41\%$), confirming the extreme imbalance of the foreground and background.
Table~\ref{tab:dataset} summarizes the key dataset properties.
A dedicated thin-stroke characterization
(Section~\ref{sec:test_splits}) found that the five thinnest-stroke
images have a mean stroke width of $11.3 \pm 0.9$\,px (measured by the
distance transform on the skeletonized masks), compared with
$21.6 \pm 18.1$\,px for the seven thickest-stroke images.

\begin{table}[t]
  \centering
  \caption{Dataset summary.}
  \label{tab:dataset}
  \begin{tabular}{l r}
    \toprule
    Property & Value \\
    \midrule
    Original images          & 34 \\
    Augmented images          & 340 \\
    Total training samples    & 374 \\
    Native resolutions        & $3712{\times}2784$, $3968{\times}2232$ \\
    Stroke coverage (mean)    & 1.79\% \\
    Stroke coverage (range)   & 0.52\%--4.94\% \\
    Test images (core)        & 7 \\
    Test images (thin)        & 5 \\
    \bottomrule
  \end{tabular}
\end{table}

\paragraph{Offline augmentation.}
Ten augmented variants are generated per original image, producing
$34 + 340 = 374$ training samples.
Augmentations are applied on two levels:
\begin{itemize}[nosep,leftmargin=*]
  \item \textbf{Weak (70\%):} brightness ($0.7$ -- $1.3$), contrast
    ($0.8$ -- $1.2$), gamma ($0.7$ -- $1.4$), color temperature shift,
 Gaussian blur, Gaussian noise.
  \item \textbf{Strong (30\%):} directional glare overlays (8~angles),
    shadow-band overlays (variable width and strength), combined
    glare+shadow.
\end{itemize}
Images identified as ``small-stroke'' (IDs 22--37) receive a gentler
augmentation profile with reduced blur and noise to preserve fine
details.

\paragraph{Online augmentation.}
During training, each sample is further augmented with random horizontal
flip (50\%), rotation ($\pm 10^{\circ}$, 50\%), color jitter
(brightness $0.3$, contrast $0.3$, saturation $0.2$, hue $0.1$),
random brightness/contrast enhancement, Gaussian blur (30\%), sharpening
(30\%) and random mask erosion with a kernel $2 \times 2$ (40\%).
The mask erosion step is specifically designed to teach the model to
preserve the separation between thin adjacent strokes.

\subsection{Architecture}
\label{sec:arch}

The segmentation model is DeepLabV3~\cite{chen2017rethinking} with a
MobileNetV3-Large~\cite{howard2019searching} backbone, loaded with
ImageNet-pretrained weights (\texttt{weights="DEFAULT"} in Torchvision).
The final classification head is replaced with a $1 \times 1$ convolution
mapping 256 channels to 2 classes (background, stroke), and the auxiliary
classifier is removed.
The model has approximately 11\,M trainable parameters.
All layers are fine-tuned; no parameters are frozen.
DeepLabV3-MobileNetV3 was chosen deliberately for three reasons:
(i)~its lightweight footprint ($\sim$11\,M parameters) matches the
deployment target of a real-time scanner on consumer GPUs;
(ii)~the atrous spatial pyramid pooling (ASPP) module provides
multi-scale context without a heavy decoder, creating a
stress test for thin-stroke preservation;
and (iii)~fixing the architecture across all runs isolates the
effect of the loss function from confounding architectural
differences, which is the central variable of this study.
Input images are resized (stretched, without aspect-ratio preservation)
to the target training resolution and normalized with
ImageNet statistics (mean $= [0.485, 0.456, 0.406]$,
std $= [0.229, 0.224, 0.225]$).

\subsection{Loss Functions}
\label{sec:losses}

Five loss functions are compared:

\paragraph{Cross-Entropy (CE).}
The standard pixel-wise cross-entropy loss, applied without class
weighting.

\paragraph{Focal Loss.}
Following Lin et al.~\cite{lin2017focal}:
\begin{equation}
  \mathcal{L}_\text{focal} = -\alpha\,(1 - p_t)^{\gamma}\,\log p_t
  \label{eq:focal}
\end{equation}
with $\alpha = 0.25$ and $\gamma = 2.0$.

\paragraph{Dice Loss.}
Following Milletari et al.~\cite{milletari2016vnet}:
\begin{equation}
  \mathcal{L}_\text{dice} = 1 - \frac{2\sum_i p_i g_i + \epsilon}
       {\sum_i p_i + \sum_i g_i + \epsilon}
  \label{eq:dice}
\end{equation}
where $p_i$ is the predicted softmax probability for the stroke class in
pixel $i$, $g_i$ is the ground-truth label, and $\epsilon = 1$.

\paragraph{Dice + Focal.}
A weighted combination:
$\mathcal{L} = 0.6\,\mathcal{L}_\text{dice} + 0.4\,\mathcal{L}_\text{focal}$.

\paragraph{Tversky Loss.}
Following Salehi et al.~\cite{salehi2017tversky}:
\begin{equation}
  \mathcal{L}_\text{tversky} = 1 - \frac{\sum_i p_i g_i + \epsilon}
       {\sum_i p_i g_i + \alpha \sum_i p_i (1-g_i)
        + \beta \sum_i (1-p_i) g_i + \epsilon}
  \label{eq:tversky}
\end{equation}
with $\alpha = 0.3$, $\beta = 0.7$, biasing toward recall.

\subsection{Evaluation Metrics}
\label{sec:metrics}

\paragraph{Region-based.}
\fone{} (Dice coefficient) and \iou{} (Jaccard index) measure
spatial overlap between the predicted and ground-truth stroke masks:
\begin{equation}
  \text{F1} = \frac{2\,\mathrm{TP}}{2\,\mathrm{TP}+\mathrm{FP}+\mathrm{FN}},
  \qquad
  \text{IoU} = \frac{\mathrm{TP}}{\mathrm{TP}+\mathrm{FP}+\mathrm{FN}},
  \label{eq:f1_iou}
\end{equation}
where TP, FP, and FN are counted at the pixel level by binarising both
predicted and ground-truth masks at a fixed threshold of~127.

\paragraph{Boundary F1 (BF1).}
Contour pixels are extracted through the morphological gradient
$\mathbf{C} = \delta_{3\times3}(\mathbf{M}) - \varepsilon_{3\times3}(\mathbf{M})$,
where $\delta$ and $\varepsilon$ denote dilation and erosion with a
square kernel $3 \times 3$.
A predicted contour pixel~$p$ is a \emph{true-positive match} if any
ground-truth contour pixel lies within Chebyshev distance~$\tau$, and
vice versa.  The tolerance is resolution-scaled:
\begin{equation}
  \tau = \max\!\bigl(1,\;\bigl\lfloor 2\,\max(H,W)/1536 \bigr\rceil\bigr)
  \;\text{pixels},
  \label{eq:tol}
\end{equation}
yielding $\tau=1$ at $1024\times768$ and $\tau\approx5$ at native
resolution.
The base tolerance of 2~pixels at the training resolution ($1536{\times}1152$)
corresponds to approximately 0.13\% of the image width, which is
comparable to the 1--2~pixel annotation uncertainty at thin-stroke
boundaries; linear scaling ensures that evaluations at different
resolutions remain comparable without inflating scores.
Boundary precision and recall quantify the proportions of correctly matched contour pixels among the predicted and ground-truth sets, respectively, while BF1 denotes the harmonic mean of these two measures~\cite{csurka2013}.

\paragraph{Boundary IoU (B-IoU).}
Following Cheng et al.~\cite{cheng2021boundary}, the IoU computation is
restricted to a boundary band of width
$d = 0.02 \cdot \sqrt{H^2 + W^2}$ (2\% of the image diagonal).
The band is efficiently computed via the Euclidean distance transform:
$G_d = \{x : \mathrm{DT}_{\bar{G}}(x) \leq d\}$, where $\bar{G}$ is
the complement of the ground-truth mask.  Predicted pixels outside the
band are ignored, ensuring that the metric is sensitive to contour placement
rather than interior fill.

\subsection{Test Splits}
\label{sec:test_splits}

Twelve images are held out for testing (never seen during training or
validation).
They are partitioned into two groups based on a
stroke-width characterization of the full dataset (computed once via the
distance transform on skeletonized masks; the partition is a fixed list
of image IDs, not a data-dependent threshold):
\begin{itemize}[nosep,leftmargin=*]
  \item \textbf{Core} (7~images: IDs 3, 13--17, 28):
    mean stroke fraction $2.41\% \pm 1.11\%$,
    mean stroke width $21.6 \pm 18.1$\,px.
  \item \textbf{Thin} (5~images: IDs 22, 24, 27, 33, 36):
    mean stroke fraction $1.14\% \pm 0.41\%$,
    mean stroke width $11.3 \pm 0.9$\,px.
\end{itemize}
The remaining 22 images and their 220 augmented variants form the
training pool, which is split 80/20 (by alphabetical filename order)
into training and validation sets.
Augmented variants of test images are automatically excluded from the training data. Specifically, any filename that begins with a test image identifier (e.g., \texttt{image\_3\_aug*}) is removed via prefix-based matching prior to the training procedure.

\section{Experiments}
\label{sec:experiments}

\subsection{Loss Ablation}
\label{sec:loss_ablation}

Five loss functions (Section~\ref{sec:losses}) are compared at a fixed
training resolution of $1536 \times 1152$ (the higher of the two
resolutions studied).
Each configuration is trained three times with seeds
$\{42, 123, 7\}$, resulting in $5 \times 3 = 15$ models.

\paragraph{Training details.}
All models use the AdamW optimizer~\cite{loshchilov2019decoupled} with
learning rate $1 \times 10^{-4}$, weight decay $1 \times 10^{-4}$,
cosine-annealing schedule with a linear warmup of 5-epochs, batch size 2,
and a maximum of 150~epochs with early stopping (patience 15, monitoring
validation loss).
Training is performed on a single NVIDIA RTX~3080 (10\,GB) with
automatic mixed precision (AMP).

\subsection{Resolution Study}
\label{sec:res_study}

To isolate the effect of input resolution on thin-structure preservation,
the best-performing loss (Dice+Focal) is trained at two resolutions.
 Loss ablation (Section~\ref{sec:loss_ablation}) is conducted at a
fixed resolution so that loss conclusions hold under a fixed compute
budget; resolution is an orthogonal lever investigated here:
\begin{itemize}[nosep,leftmargin=*]
  \item $1024 \times 768$ (the Torchvision default for DeepLabV3 demos),
  \item $1536 \times 1152$ (approximately $2\times$ in each dimension).
\end{itemize}
All other hyperparameters are identical to the loss ablation.
Three seeds per resolution yield $2 \times 3 = 6$ models; the
$1536 \times 1152$ models overlap with the Dice+Focal runs in the
loss ablation.

\subsection{Classical Baseline}
\label{sec:baseline}

Three baseline methods based on non-learning are evaluated on the identical 12-image test set at the \emph{original} spatial resolution of each image (i.e. without any resizing) to ensure a fair comparison:
\begin{itemize}[nosep,leftmargin=*]
  \item \textbf{Adaptive thresholding} using OpenCV’s
    \texttt{adaptiveThreshold} implementation (Gaussian method, block size $51$, and offset parameter $C = 15$).
  \item \textbf{Otsu thresholding}, corresponding to a global binarization of the grayscale image through Otsu's between-class variance maximization criterion.
  \item \textbf{Sauvola thresholding}~\cite{sauvola2000adaptive}, i.e.\ local adaptive thresholding with
    \[
      T(x,y) = \mu(x,y)\Bigl[1 + k\bigl(\sigma(x,y)/R - 1\bigr)\Bigr],
    \]
    using a window size of $51$ and a parameter $k = 0.2$.
    Sauvola thresholding is widely adopted as a reference method for document-image binarization.
\end{itemize}
The predictions produced by deep learning models are rescaled to the original image resolution via nearest-neighbor interpolation prior to the metric computation, ensuring that all methods are evaluated at the native pixel resolution.

\subsection{Metrics}
\label{sec:metrics_exp}

Each model is evaluated in the 12 test images that have not been tested using four
metrics: \fone{}, \iou{}, BF1, and B-IoU
(Section~\ref{sec:metrics}).
The metrics are computed per image and then averaged.
The results are reported as mean $\pm$ standard deviation in three
seeds.
Statistical significance is assessed with the Wilcoxon signed-rank test
on per-image scores (12 paired samples, averaged between seeds), with
Bonferroni correction ($\alpha_\text{corr} = 0.05 / 10 = 0.005$ for
10 pairwise comparisons).

\paragraph{Scope of the loss comparison.}
We deliberately compare widely-used \emph{region/overlap} losses (CE,
Focal, Dice, Tversky, and their combinations) rather than
boundary-optimized losses such as
InverseForm~\cite{borse2021inverseform} or
clDice~\cite{shit2021cldice}.
Our contribution is the \emph{boundary-metric evaluation protocol}
itself, not a new loss function.
Because the protocol is architecture- and loss-agnostic, it can be
applied to boundary-aware losses in future work
(Section~\ref{sec:conclusion}).

\section{Results}
\label{sec:results}

\subsection{Loss Ablation}
\label{sec:results_loss}

Table~\ref{tab:loss_study} summarizes the region and boundary metrics for all
five loss functions.
Dice-family losses (Dice, Dice+Focal, Tversky) uniformly outperform 
distribution-based losses (CE, Focal) by a wide margin: Tversky achieves
an \fone{} of 0.663 versus 0.438 for CE -- a gain of more than 20
percentage points.
The difference is highly significant (Wilcoxon $p < 0.001$;
Table~\ref{tab:stat_sig}).
Among Dice-family losses, the differences are not statistically
significant ($p > 0.15$ for all pairwise comparisons), suggesting
that the primary benefit is simply switching away from
a distribution-based objective rather than tuning a specific
overlap-based variant.

\begin{table}[t]
  \centering
  \caption{Loss ablation at $1536{\times}1152$ on 12 held-out test
  images (mean $\pm$ std across three seeds).  Classical baselines
  (Table~\ref{tab:baseline}) are evaluated at native resolution.
  Dice-family losses improve \fone{} by more than 20
  points over CE.}
  \label{tab:loss_study}
  \resizebox{\columnwidth}{!}{\begin{tabular}{l S[table-format=1.3] @{${\pm}$} S[table-format=1.3] S[table-format=1.3] @{${\pm}$} S[table-format=1.3] S[table-format=1.3] @{${\pm}$} S[table-format=1.3] S[table-format=1.3] @{${\pm}$} S[table-format=1.3]}
\toprule
Loss & \multicolumn{2}{c}{F1} & \multicolumn{2}{c}{IoU} & \multicolumn{2}{c}{BF1} & \multicolumn{2}{c}{B-IoU} \\
\midrule
CE & 0.438 & 0.011 & 0.288 & 0.009 & 0.594 & 0.007 & 0.291 & 0.010 \\
Focal & 0.430 & 0.012 & 0.283 & 0.009 & 0.589 & 0.013 & 0.285 & 0.009 \\
Dice & 0.650 & 0.004 & 0.486 & 0.005 & 0.674 & 0.002 & 0.490 & 0.005 \\
Dice+Focal & 0.657 & 0.005 & 0.494 & 0.006 & 0.676 & 0.002 & 0.497 & 0.005 \\
Tversky & 0.663 & 0.002 & 0.500 & 0.002 & 0.646 & 0.003 & 0.505 & 0.002 \\
\midrule
Adaptive (baseline) & 0.761 & 0.112 & 0.626 & 0.132 & 0.853 & 0.231 & 0.628 & 0.130 \\
Sauvola (baseline)  & 0.787 & 0.112 & 0.660 & 0.131 & 0.860 & 0.229 & 0.663 & 0.131 \\
\bottomrule
\end{tabular}
}
\end{table}

\begin{figure}[t]
  \centering
  \includegraphics[width=\columnwidth]{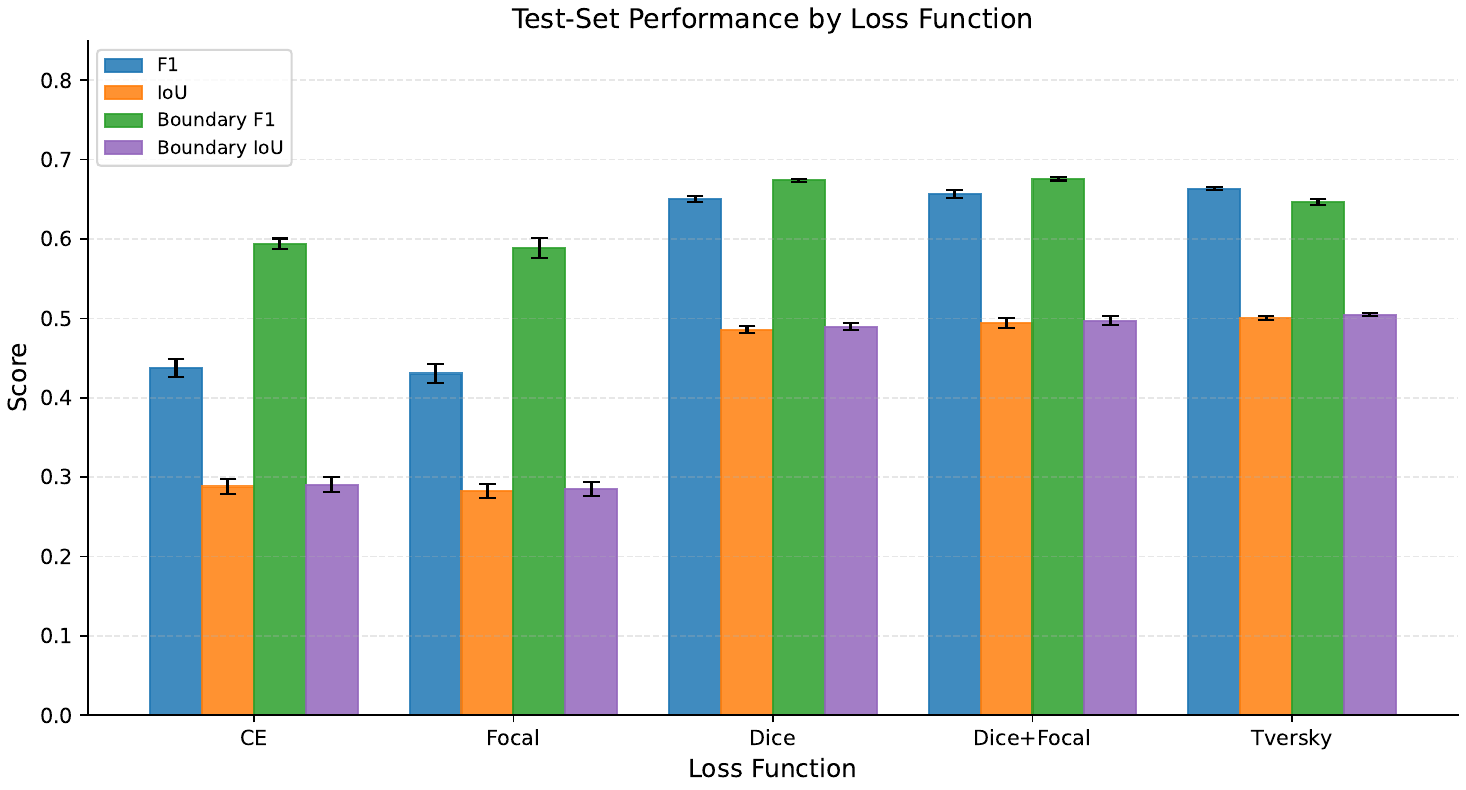}
  \caption{Region and boundary metrics per loss function.
  Dice-based objectives form a clearly separated, higher-performance
  cluster.}
  \label{fig:loss_bars}
\end{figure}

Figure~\ref{fig:loss_bars} visualizes these differences.  CE and Focal
cluster together at approximately 0.43 \fone{}, while the three
 losses of the Dice-family cluster near 0.66, confirming that the choice 
of loss is the dominant factor in this study.

\subsection{Core vs.\ Thin Subsets}
\label{sec:results_core_thin}

Table~\ref{tab:core_thin} breaks the performance into seven 
\emph{core} images and five \emph{thin-stroke} images.  All losses
degrade in the thin subset, but CE and Focal suffer a larger
drop (gap $\approx 0.10$) than losses in the Dice-family (gap $\approx 0.06$).
Tversky achieves the highest thin-subset \fone{} (0.628) with the
smallest gap, consistent with its recall-biased design.

\begin{table}[t]
  \centering
  \caption{Core vs.\ thin-subset \fone{}.  Dice-family losses halve the
  performance gap between subsets compared with CE/Focal, indicating
  better generalization to thin strokes.}
  \label{tab:core_thin}
  \resizebox{\columnwidth}{!}{\begin{tabular}{l S[table-format=1.3] @{${\pm}$} S[table-format=1.3] S[table-format=1.3] @{${\pm}$} S[table-format=1.3] S[table-format=1.3]}
\toprule
Loss & \multicolumn{2}{c}{Core F1} & \multicolumn{2}{c}{Thin F1} & {Gap} \\
\midrule
CE & 0.480 & 0.010 & 0.378 & 0.013 & 0.102 \\
Focal & 0.476 & 0.010 & 0.367 & 0.015 & 0.109 \\
Dice & 0.674 & 0.006 & 0.618 & 0.003 & 0.056 \\
Dice+Focal & 0.685 & 0.009 & 0.618 & 0.003 & 0.067 \\
Tversky & 0.689 & 0.003 & 0.628 & 0.001 & 0.061 \\
\bottomrule
\end{tabular}
}
\end{table}

\begin{figure}[t]
  \centering
  \includegraphics[width=\columnwidth]{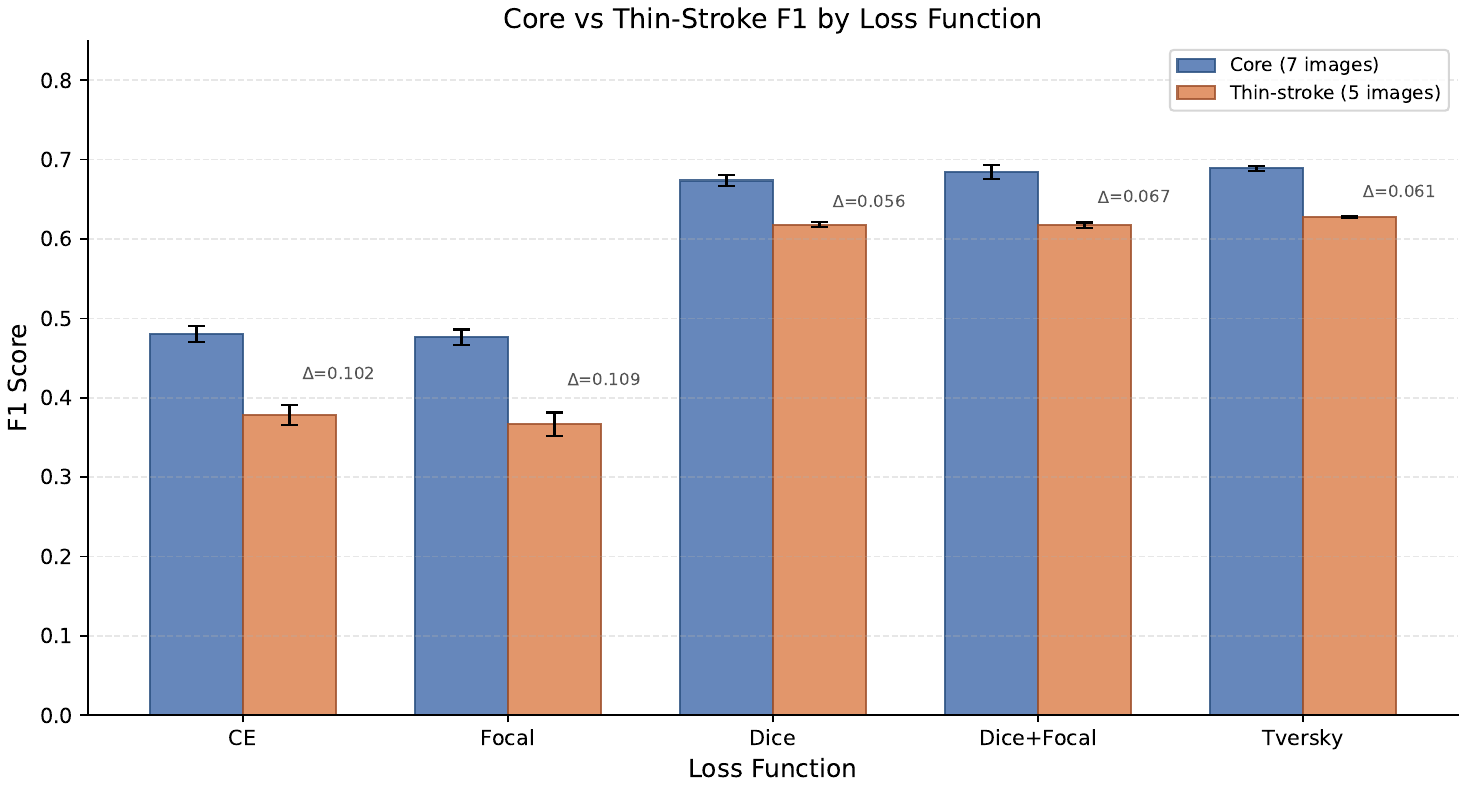}
  \caption{Core vs.\ thin \fone{} per loss.  Dice-family losses
  narrow the gap, with Tversky showing the most balanced performance.}
  \label{fig:core_thin}
\end{figure}

Figure~\ref{fig:core_thin} presents these results graphically:
the gap between the two clusters narrows substantially for overlap-based
losses, with Tversky offering the most balanced core/thin trade-off.

\begin{figure}[t]
  \centering
  \includegraphics[width=\columnwidth]{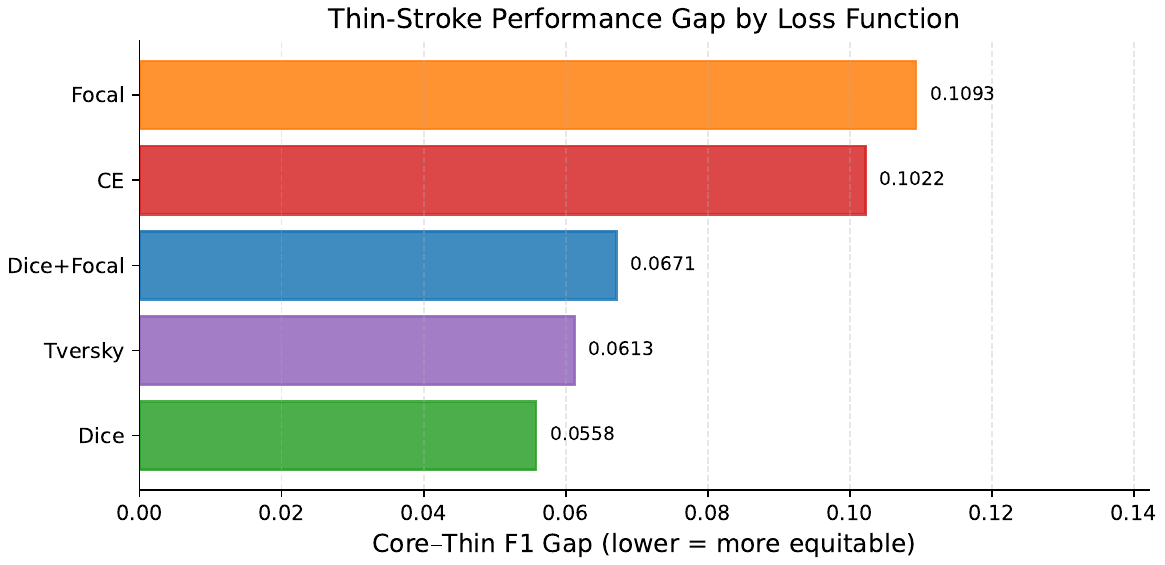}
  \caption{Core--thin \fone{} gap per loss.  A smaller gap indicates
  more equitable treatment of thin strokes.}
  \label{fig:core_thin_gap}
\end{figure}

Figure~\ref{fig:core_thin_gap} plots the absolute gap between the core and
thin \fone{}.  CE and Focal exhibit gaps exceeding 0.10, while all
Dice-family losses remain below 0.07.

\subsection{Statistical Significance and Effect Sizes}
\label{sec:results_stat}

Table~\ref{tab:stat_sig} reports pairwise statistical tests with both
mean and median effect sizes per-image.
All CE/Focal vs.\ Dice-family comparisons are significant at
$p < 0.001$ after Bonferroni correction, with a median per-image gain
of $+0.225$ \fone{} points.
The mean per-image improvement from CE to Dice is
$+0.213 \pm 0.067$ \fone{}, and from CE to Tversky
$+0.226 \pm 0.085$ \fone{} ($p < 0.001$ in both cases),
confirming that the effect is both statistically significant and
practically large.
Intra-Dice-family comparisons are not significant ($p > 0.15$),
confirming that the dominant effect is the switch from distribution-based
to overlap-based objectives, not the specific variant.

\begin{table}[t]
  \centering
  \caption{Pairwise comparisons with mean and median per-image effect
  sizes (seed-averaged).  *** = $p < 0.001$ (Wilcoxon, Bonferroni-corrected,
  $\alpha = 0.005$); n.s.\ = not significant.}
  \label{tab:stat_sig}
  \resizebox{\columnwidth}{!}{\begin{tabular}{l S[table-format=+1.4] S[table-format=+1.4] S[table-format=1.4] l}
\toprule
Comparison & {$\Delta$F1 (mean)} & {$\Delta$F1 (median)} & {$p$ (Wilcoxon)} & {Sig.} \\
\midrule
CE vs.~Focal & +0.0074 & +0.0063 & 0.2661 & n.s. \\
CE vs.~Dice & -0.2126 & -0.2253 & 0.0005 & *** \\
CE vs.~Dice+Focal & -0.2189 & -0.2214 & 0.0005 & *** \\
CE vs.~Tversky & -0.2256 & -0.2369 & 0.0005 & *** \\
Focal vs.~Dice & -0.2199 & -0.2346 & 0.0005 & *** \\
Focal vs.~Dice+Focal & -0.2262 & -0.2392 & 0.0005 & *** \\
Focal vs.~Tversky & -0.2330 & -0.2453 & 0.0005 & *** \\
Dice vs.~Dice+Focal & -0.0063 & +0.0000 & 0.8501 & n.s. \\
Dice vs.~Tversky & -0.0131 & -0.0055 & 0.1514 & n.s. \\
Dice+Focal vs.~Tversky & -0.0068 & -0.0086 & 0.3804 & n.s. \\
\bottomrule
\end{tabular}
}
\end{table}

\subsection{Resolution Study}
\label{sec:results_res}

Table~\ref{tab:resolution} shows that increasing resolution from
$1024{\times}768$ to $1536{\times}1152$ increases Dice+Focal \fone{} from
0.530 to 0.657 (+0.127) and BF1 from 0.491 to 0.676 (+0.185), a
greater relative gain in boundary quality.
Thin strokes span fewer pixels at reduced resolution, making them
harder to recover.

\begin{table}[t]
  \centering
  \caption{Resolution ablation (Dice+Focal, three seeds).  Doubling
  resolution yields a +13-point \fone{} gain and an even larger
  boundary-metric improvement.}
  \label{tab:resolution}
  \resizebox{\columnwidth}{!}{\begin{tabular}{l S[table-format=1.3] @{${\pm}$} S[table-format=1.3] S[table-format=1.3] @{${\pm}$} S[table-format=1.3] S[table-format=1.3] @{${\pm}$} S[table-format=1.3] S[table-format=1.3] @{${\pm}$} S[table-format=1.3]}
\toprule
Resolution & \multicolumn{2}{c}{F1} & \multicolumn{2}{c}{IoU} & \multicolumn{2}{c}{BF1} & \multicolumn{2}{c}{B-IoU} \\
\midrule
1024$\times$768 & 0.530 & 0.005 & 0.369 & 0.006 & 0.491 & 0.006 & 0.372 & 0.006 \\
1536$\times$1152 & 0.657 & 0.005 & 0.494 & 0.006 & 0.676 & 0.002 & 0.497 & 0.005 \\
\bottomrule
\end{tabular}
}
\end{table}

\begin{figure}[t]
  \centering
  \includegraphics[width=\columnwidth]{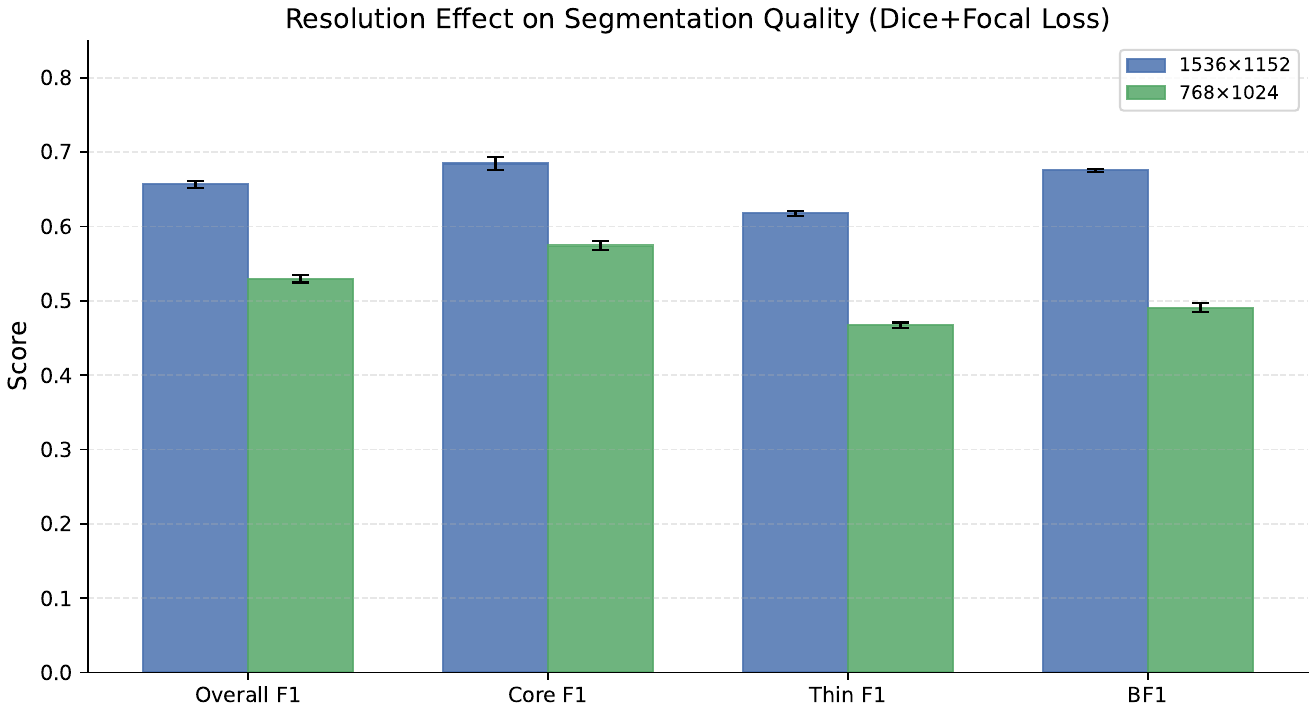}
  \caption{Resolution comparison for Dice+Focal.  Boundary metrics
  benefit more than region metrics from higher resolution.}
  \label{fig:resolution}
\end{figure}

\subsection{Classical Baseline}
\label{sec:results_baseline}

Table~\ref{tab:baseline} compares three classical thresholding methods.
Sauvola thresholding, a standard baseline for
document--images, achieves the highest classical mean \fone{} (0.787), closely followed by adaptive Gaussian thresholding (0.761).
However, both local methods share considerable fragility:
the worst-case \fone{} drops to 0.452 (Sauvola) and 0.458
(adaptive) --below the Tversky minimum of 0.565.
Otsu thresholding is included as a sanity baseline; its near-zero
\fone{} (0.059) confirms that global binarization cannot cope with
uneven whiteboard backgrounds and is not a competitive reference.

\begin{table}[t]
  \centering
  \caption{Classical baselines at original resolution.
  Sauvola and adaptive thresholding achieve high means but show the
  widest per-image spread in the study; Otsu fails entirely.}
  \label{tab:baseline}
  \resizebox{\columnwidth}{!}{\begin{tabular}{l S[table-format=1.3] @{${\pm}$} S[table-format=1.3] S[table-format=1.3] @{${\pm}$} S[table-format=1.3] S[table-format=1.3] @{${\pm}$} S[table-format=1.3] S[table-format=1.3] @{${\pm}$} S[table-format=1.3]}
\toprule
Method & \multicolumn{2}{c}{F1} & \multicolumn{2}{c}{IoU} & \multicolumn{2}{c}{BF1} & \multicolumn{2}{c}{B-IoU} \\
\midrule
Adaptive & 0.761 & 0.112 & 0.626 & 0.132 & 0.853 & 0.231 & 0.628 & 0.130 \\
Otsu & 0.059 & 0.027 & 0.031 & 0.014 & 0.202 & 0.067 & 0.055 & 0.021 \\
Sauvola & 0.787 & 0.112 & 0.660 & 0.131 & 0.860 & 0.229 & 0.663 & 0.131 \\
\bottomrule
\end{tabular}
}
\end{table}

\paragraph{When to prefer each approach.}
For an application that tolerates occasional poor frames (e.g., batch
archiving with manual review), Sauvola thresholding is the simplest
choice: no GPU, no training, and a strong mean.
For an application that requires consistent quality across all images
(e.g., real-time note capture), a learned model is preferable: Tversky
never drops below 0.565 \fone{}, whereas the best classical method
fails on low-contrast boards (0.452).  The gap narrows further when
the learned model operates at a higher resolution
(Section~\ref{sec:results_res}).
The mean \fone{} can be dominated by easy and high-contrast images;
boundary metrics more directly evaluate thin-stroke fidelity, which is
the primary concern in a scanner pipeline.

\begin{figure*}[t]
  \centering
  \includegraphics[width=\textwidth]{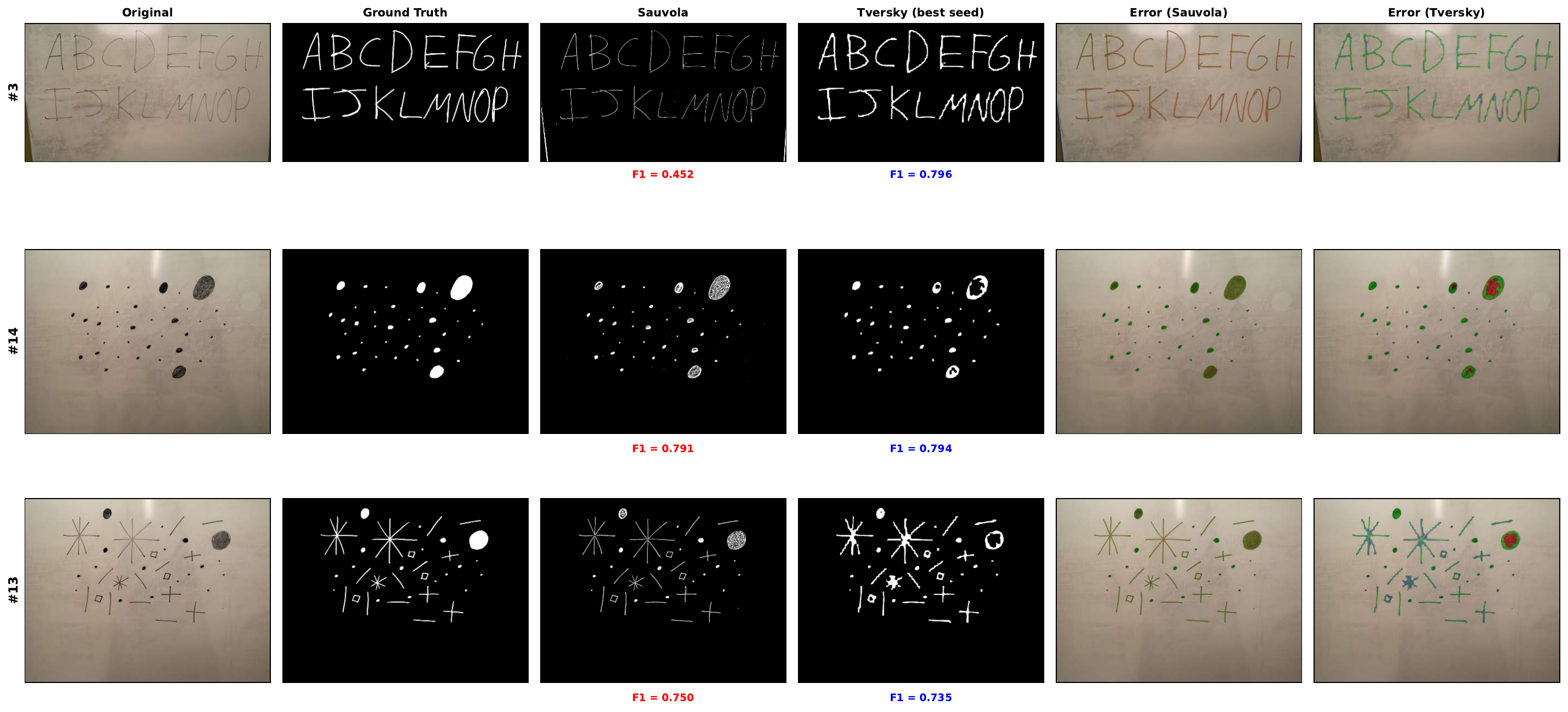}
  \caption{Classical baseline failure cases.  Rows are the three images
  where Sauvola binarization scores lowest.  Columns: original image,
  ground truth, Sauvola prediction, Tversky prediction, and
  error overlays (green = TP, red = FN, blue = FP).  The deep model
  provides more consistent segmentation on challenging boards.}
  \label{fig:baseline_failure}
\end{figure*}

Figure~\ref{fig:baseline_failure} shows the three images in which the
Sauvola baseline performs the worst. In image~\#3 -- a low-contrast board
with uneven lighting---Sauvola drops to \fone{} = 0.452,
while Tversky achieves 0.796.  The error overlays make the failure mode
concrete: local binarization hallucinates false-positive strokes in
shadow regions (blue) while simultaneously missing faint strokes (red).

\subsection{Robustness Analysis}
\label{sec:results_robustness}

Table~\ref{tab:robustness} goes beyond the mean $\pm$ std and reports
distributional robustness statistics for each method: median, IQR,
worst-case (min), best-case (max) and paired wins against the
strongest classical baseline (Sauvola).

\begin{table}[t]
  \centering
  \caption{Robustness profile (per-image \fone{}, seed-averaged).
  Deep models exhibit tighter IQR and higher worst-case \fone{} than
  classical baselines despite a lower mean.
  ``Wins/12'' = number of test images where the method's \fone{} exceeds the best deep model's (for classical rows) or Sauvola's (for deep rows).}
  \label{tab:robustness}
  \resizebox{\columnwidth}{!}{\begin{tabular}{l S[table-format=1.3] S[table-format=1.3] S[table-format=1.3] S[table-format=1.3] S[table-format=1.3] r}
\toprule
Method & {Mean} & {Median} & {IQR} & {Min} & {Max} & {Wins/12} \\
\midrule
CE & 0.438 & 0.413 & 0.157 & 0.230 & 0.667 & 1/12 \\
Focal & 0.430 & 0.398 & 0.183 & 0.223 & 0.684 & 1/12 \\
Dice & 0.650 & 0.653 & 0.054 & 0.542 & 0.791 & 1/12 \\
Dice+Focal & 0.657 & 0.649 & 0.072 & 0.528 & 0.789 & 2/12 \\
Tversky & 0.663 & 0.651 & 0.066 & 0.565 & 0.795 & 1/12 \\
\midrule
Sauvola & 0.787 & 0.800 & 0.081 & 0.452 & 0.893 & 10/12 \\
Adaptive & 0.761 & 0.780 & 0.131 & 0.458 & 0.884 & 9/12 \\
\bottomrule
\end{tabular}
}
\end{table}

Key observations:
\begin{itemize}[nosep,leftmargin=*]
  \item Sauvola wins on 10 of 12 images in absolute \fone{}, but
    Dice-family losses beat it on the low-contrast boards where
    local binarization struggles most.
  \item Despite losing on most images, the deep models' IQR is
    narrower (e.g., 0.054 for Dice vs.\ 0.081 for Sauvola),
    meaning they are more predictable.
  \item Tversky's worst-case \fone{} (0.565) is the highest among all
    methods including Sauvola (0.452), making it the safest choice
    for failure-sensitive deployments.
  \item CE and Focal have IQRs far wider than Sauvola ($\sim$0.16
    --0.18 vs.\ 0.081), indicating that under extreme imbalance,
    CE-family losses are not only worse on average, but also less stable.
\end{itemize}

\begin{figure}[t]
  \centering
  \includegraphics[width=\columnwidth]{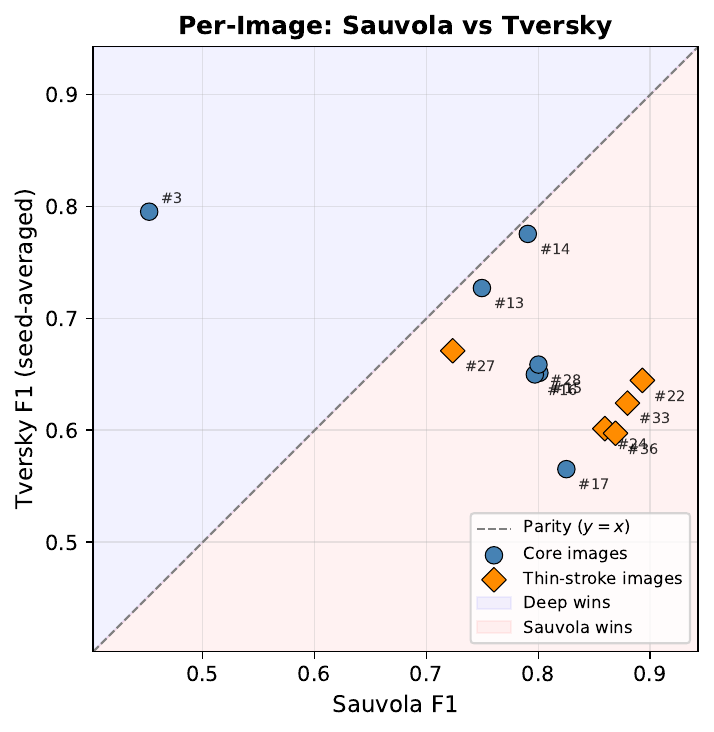}
  \caption{Per-image \fone{}: Sauvola (x-axis) vs.\
  Tversky (y-axis).  Points below the diagonal are images where Tversky
  outperforms the baseline.  Diamonds = thin-stroke images.}
  \label{fig:scatter}
\end{figure}

Figure~\ref{fig:scatter} reinforces this pattern: most images cluster
above the parity line (Sauvola wins), but a distinct group of
low-contrast boards falls well below it.
In these images, the advantage of the learned model exceeds 10~F1 points.

\subsection{Per-Image Analysis}
\label{sec:results_per_image}

\begin{figure}[t]
  \centering
  \includegraphics[width=\columnwidth]{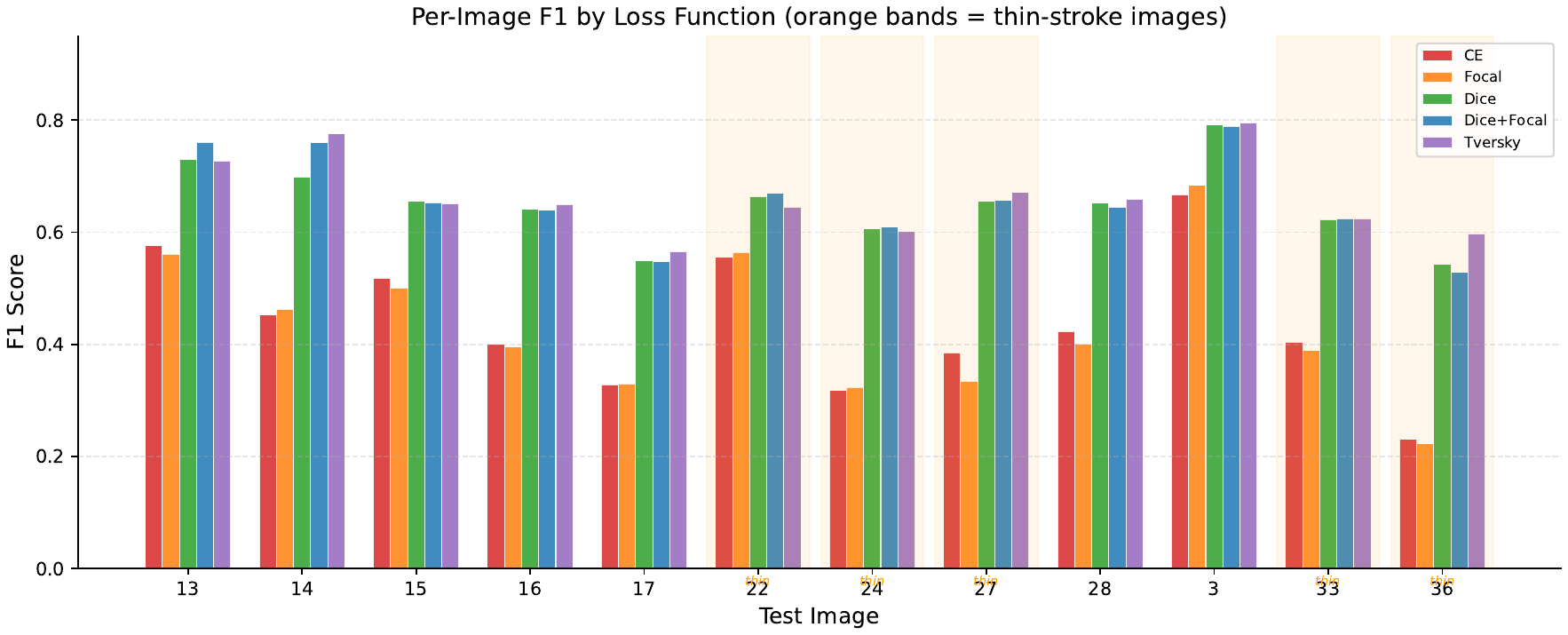}
  \caption{Per-image \fone{} across losses.  Dice-family losses are
  more consistent; CE/Focal show large per-image variation.}
  \label{fig:per_image}
\end{figure}

Figure~\ref{fig:per_image} shows per-image \fone{} scores across all
losses.  The most difficult images are those with the thinnest strokes
or the lowest contrast.  Dice-family losses consistently narrow the gap
between easy and hard images, whereas CE and Focal exhibit bimodal
behavior, nearly succeeding on some images and completely failing
on others.

\subsection{Training Dynamics}
\label{sec:results_training}

\begin{figure*}[t]
  \centering
  \includegraphics[width=\textwidth]{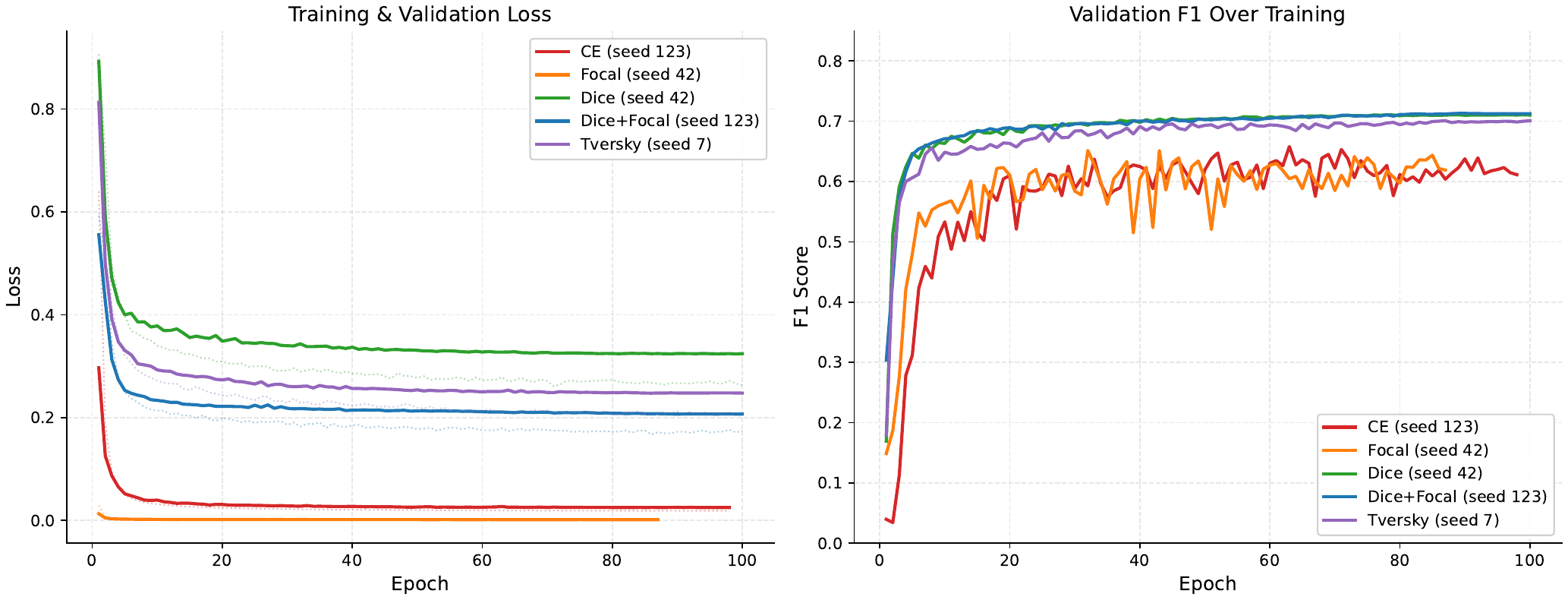}
  \caption{Training curves for each loss.  Dice-family losses show
  a gradual, steady descent; CE/Focal converge early and plateau.}
  \label{fig:training_curves}
\end{figure*}

Figure~\ref{fig:training_curves} plots the training and validation loss
curves for each loss function.  CE and Focal converge quickly but
plateau early, whereas Dice-family losses exhibit a longer, steadier
descent consistent with their more informative gradients under
class imbalance.

\subsection{Qualitative Results}
\label{sec:results_qual}

\begin{figure}[t]
  \centering
  \includegraphics[width=\columnwidth]{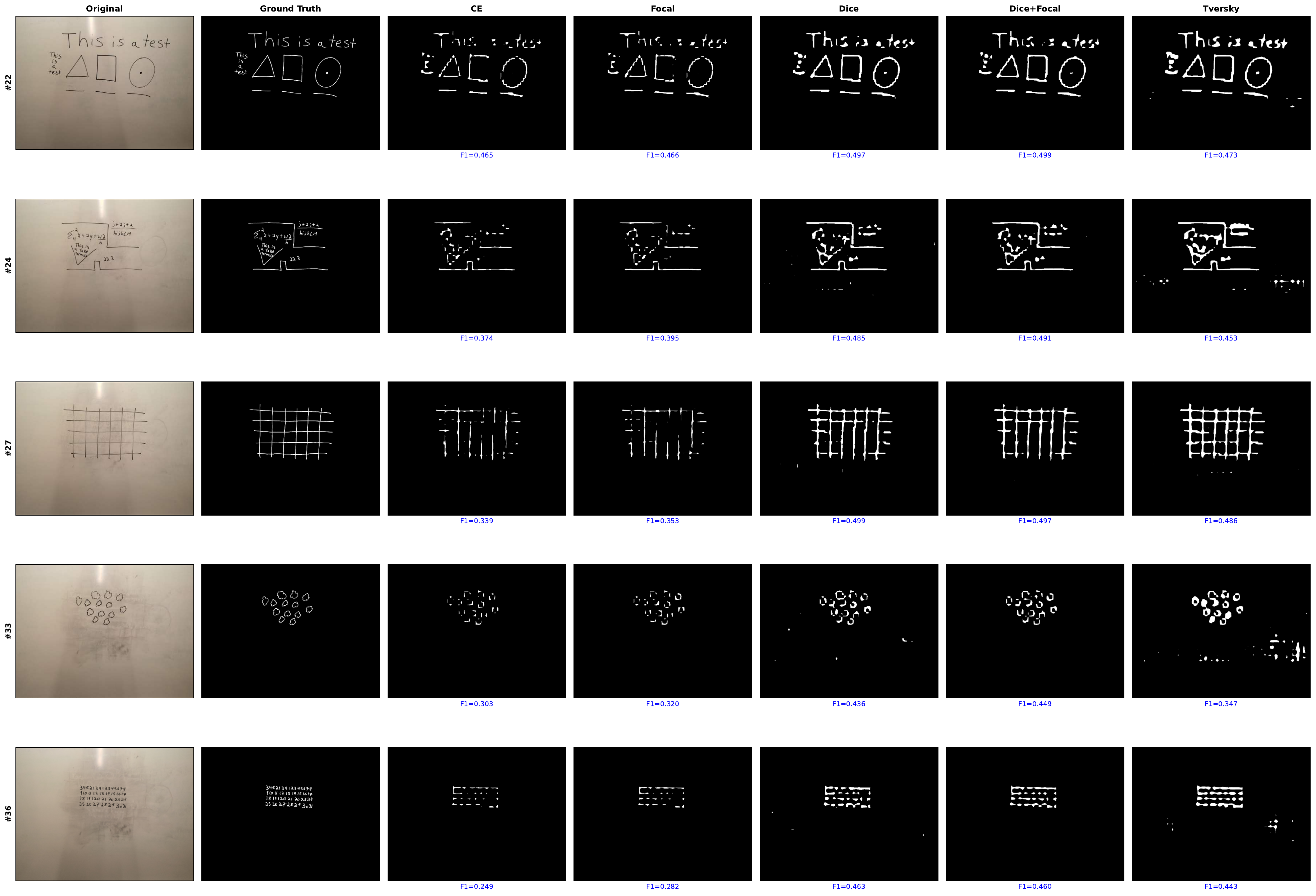}
  \caption{Qualitative predictions on thin-stroke images.  Dice-family
  losses maintain stroke continuity where CE/Focal fragment.}
  \label{fig:qual_thin}
\end{figure}

\begin{figure}[t]
  \centering
  \includegraphics[width=\columnwidth]{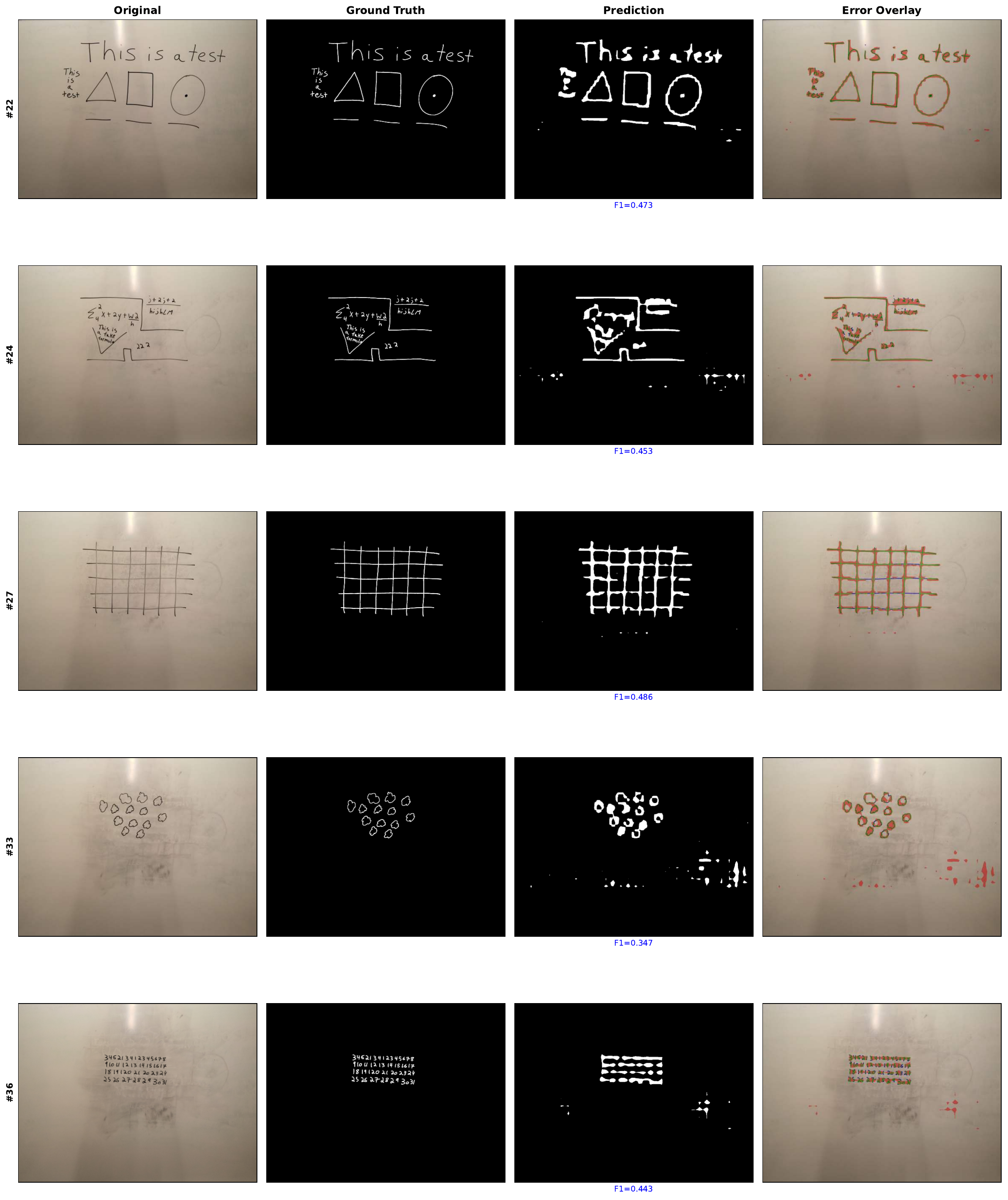}
  \caption{Error maps for thin-stroke images.  Green = true positive,
  red = false negative, blue = false positive.  CE/Focal show
  extensive false negatives on fine strokes.}
  \label{fig:error_thin}
\end{figure}

Figure~\ref{fig:qual_thin} presents predictions on thin-stroke images.
CE/Focal produce fragmented masks that miss fine details, while
Dice-family losses preserve stroke continuity.
Figure~\ref{fig:error_thin} shows error maps for the same images,
highlighting false negatives along thin strokes for CE/Focal.

\begin{figure}[t]
  \centering
  \includegraphics[width=\columnwidth]{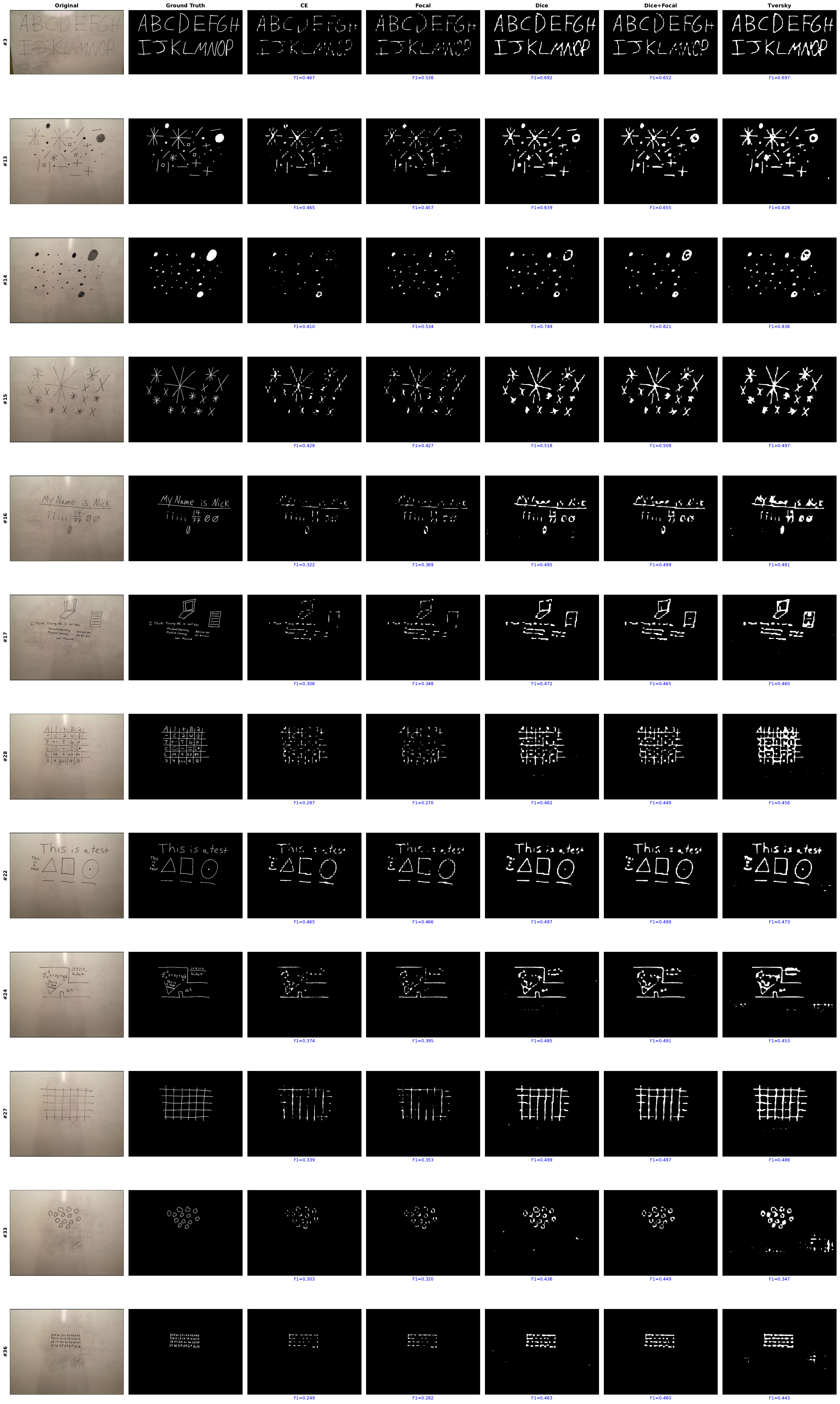}
  \caption{Qualitative predictions across the full test set.
  Performance trends observed on thin strokes generalise to
  standard images.}
  \label{fig:qual_all}
\end{figure}

\begin{figure}[t]
  \centering
  \includegraphics[width=\columnwidth,height=0.88\textheight,keepaspectratio]{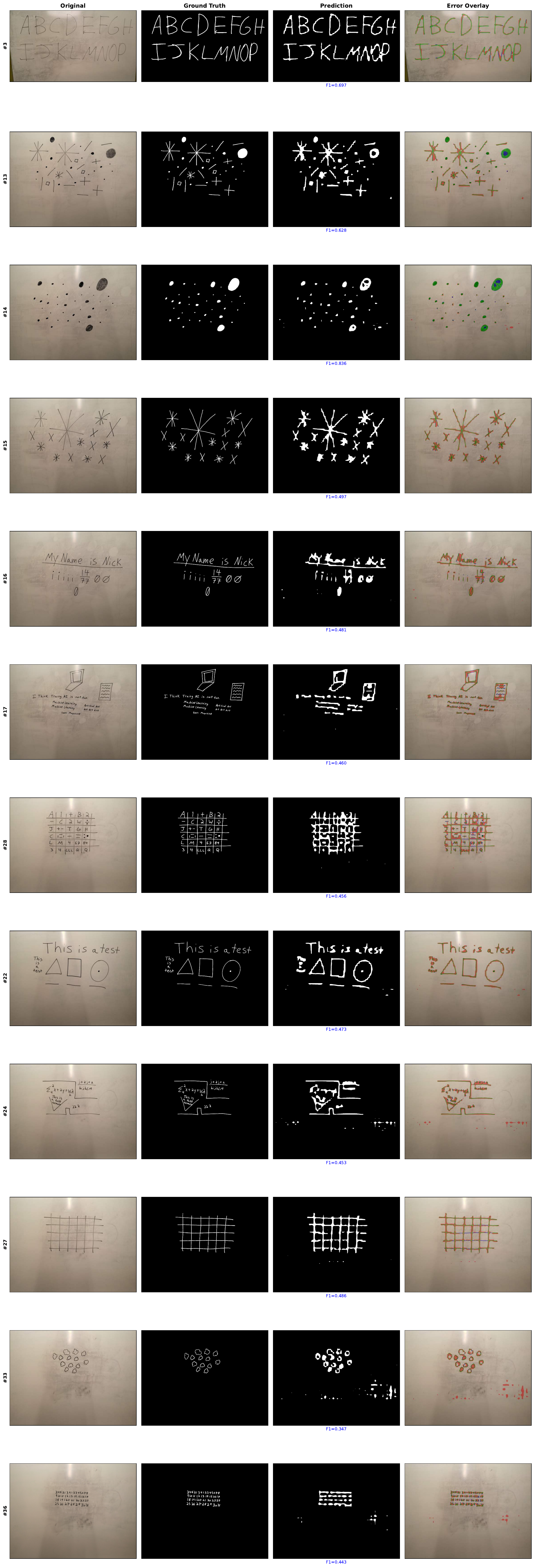}
  \caption{Error maps for the full test set.  Dice-family losses
  consistently reduce false negatives compared with CE/Focal.}
  \label{fig:error_all}
\end{figure}

Figures~\ref{fig:qual_all} and Figure~\ref{fig:error_all} show
analogous results throughout the entire test set, confirming that 
qualitative improvements extend beyond the thin subset.

\subsection{Metric Correlations}
\label{sec:results_heatmap}

\begin{figure}[t]
  \centering
  \includegraphics[width=\columnwidth]{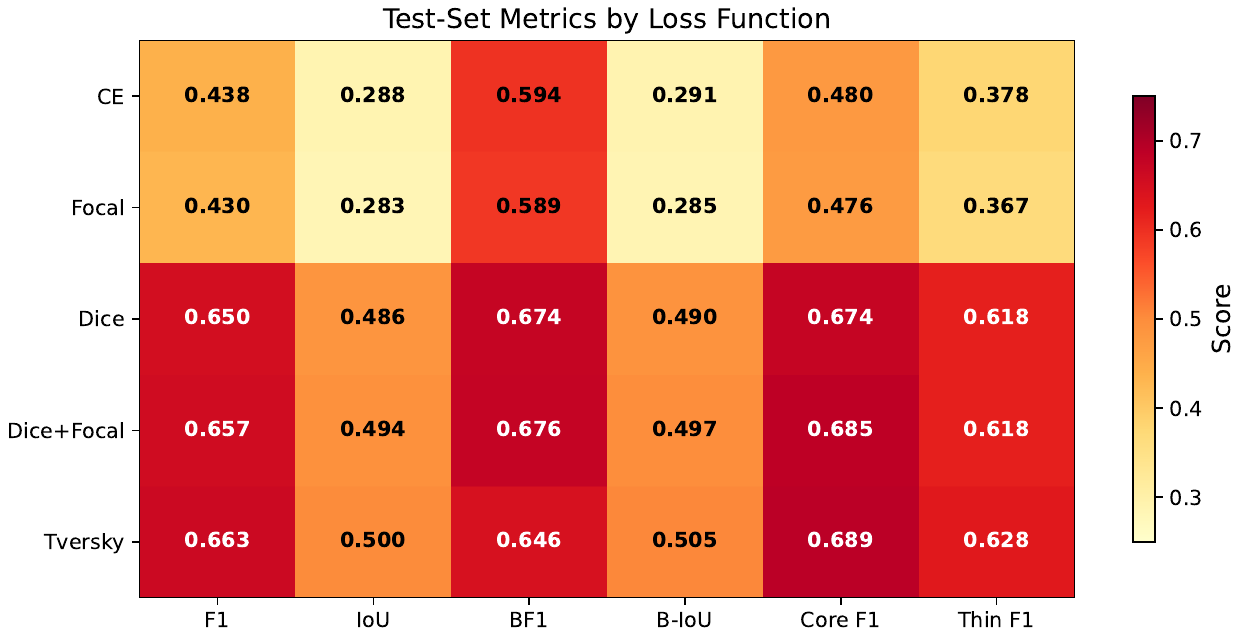}
  \caption{Metric heatmap across losses $\times$ seeds.  Boundary
  metrics differentiate models that region metrics rank similarly.}
  \label{fig:heatmap}
\end{figure}

Figure~\ref{fig:heatmap} presents a heatmap of all metrics across
losses and seeds.  BF1 and B-IoU reveal subtle differences between
Dice-family variants that region metrics alone would not surface;
for example, Dice+Focal and Dice achieve slightly higher BF1 than
Tversky despite lower \fone{}, suggesting a precision--recall
trade-off at the boundary level.

\section{Discussion}
\label{sec:discussion}

\subsection{Why Dice-Family Losses Dominate}

The 20+ point \fone{} gap between CE/Focal and Dice-family losses has
a straightforward explanation: under extreme class imbalance (stroke
pixels $\approx$ 1.8\% of the image), CE is dominated by the
easy-negative background.  A model predicting ``all background''
achieves $>$98\% pixel accuracy, so the per-pixel CE gradient provides
almost no signal for the minority class.  Focal loss re-weights by
$(1-p)^\gamma$, but most background pixels are already confident,
leaving the effective gradient budget nearly unchanged.

Dice loss, by contrast, directly optimizes overlap and is invariant to
the ratio of foreground to background pixels.  Its gradient is
proportional to the current Dice coefficient, which starts near zero for
the minority class, creating a steep learning signal precisely where it
is needed.  This mechanism has been theoretically analyzed
~\cite{milletari2016vnet,yeung2022unified} and is
consistent with our empirical results.

\subsection{Tversky's Recall Bias}

The Tversky loss ($\alpha\!=\!0.3,\;\beta\!=\!0.7$) penalizes false
negatives more than false positives, which should benefit thin strokes
that are easily missed.  In our results, Tversky achieves the highest
\fone{} (0.663) and the thin-subset \fone{} (0.628), but the differences
from Dice and Dice+Focal are not statistically significant.  A possible
explanation is that the recall--precision trade-off mainly shifts
predictions near the existing boundary rather than recovering entirely
missed strokes; this is corroborated by Tversky's slightly lower BF1
(0.646 vs.\ 0.676 for Dice+Focal), indicating marginally less precise
boundaries despite better region overlap.

\subsection{Boundary Metrics Reveal Hidden Differences}

The region metrics (\fone{}, IoU) rank the Dice-family losses similarly,
but the boundary metrics expose a subtle trade-off.  Dice+Focal achieves the
highest BF1 (0.676), suggesting crisper boundaries, while Tversky leads
in \fone{} and B-IoU.  Without boundary metrics, this distinction would
be invisible.  We therefore endorse the recent recommendation of
Cheng~et~al.~\cite{cheng2021boundary} to report boundary-aware metrics
alongside region-based ones, particularly for thin-structure tasks.

\subsection{Consistency--Accuracy Trade-off}

Sauvola binarization, the strongest classical baseline, achieves a higher mean \fone{} (0.787) than every deep model, but robustness analysis (Table~\ref{tab:robustness})
reveals the cost: an IQR of 0.081 and a worst-case \fone{} as low as
0.452, compared to IQR $\leq$ 0.072 and worst-case $\geq$ 0.542 for
every Dice-family model. Nevertheless, losses from the Dice-family beat 
Sauvola in 3 of 12 test images, specifically 
low-contrast boards where local binarization fails more
catastrophically (Sauvola drops to \fone{} = 0.452 in these images).

This pattern exposes a \emph{consistency--accuracy trade-off}: the
classical pipeline converts high-resolution information into high mean
performance but provides no robustness guarantee, whereas learned models
sacrifice some mean accuracy in exchange for substantially tighter
per-image variance.  The practical implication depends on the
deployment scenario.  For batch archiving of high-quality photos, the
classical approach may suffice; for real-time capture under variable
conditions, a learned model is preferable despite its lower ceiling.

\subsection{Resolution as a Bottleneck}

The resolution study confirms that the input resolution is a critical
bottleneck: doubling each spatial dimension improves \fone{} by 13
points and BF1 by 19 points.  Thin strokes that span only 1--2 pixels
in $1024\times768$ may span 3--4 pixels in $1536\times1152$, crossing
the threshold at which the network can reliably segment them.
Importantly, the loss-function ranking established in
Section~\ref{sec:results_loss} was obtained at a constrained resolution
to keep the comparison controlled, the deployed pipeline can operate
at higher resolutions where the absolute \fone{} is substantially higher.
Resolution and loss choice are thus orthogonal levers: the former
raises the performance ceiling, the latter determines which loss
best exploits a given resolution.
Future work should explore native-resolution or multi-scale
approaches to close the remaining gap with classical baselines.

\subsection{Failure Cases}

Despite the improvements afforded by overlap-based losses, systematic
failures remain.  The most prominent failure mode occurs on extremely
low-contrast strokes photographed under heavy directional glare: in
these regions the stroke-to-background intensity difference falls below
the network's discrimination threshold, producing fragmented or entirely
missed predictions.  A secondary failure mode involves very fine strokes
($<$8\,px width at training resolution) that are eroded to sub-pixel
width after downsampling, leaving insufficient signal for any loss
function to recover.  These cases are visible in the error maps
(Figure~\ref{fig:error_thin}), where red false-negative bands
persist along the thinnest strokes even for the best-performing
Tversky model.

\subsection{Practical Implications}

For production whiteboard digitization systems operating under variable
lighting conditions and with mixed marker types, we recommend Dice or
Tversky losses over cross-entropy whenever stroke coverage falls below
approximately 5\%.  The consistency advantage of learned models---higher
worst-case \fone{} and tighter IQR---is particularly valuable in
real-time capture pipelines where manual review is impractical.
Where computational resources permit, training at the highest feasible
resolution provides a complementary and orthogonal gain.

\subsection{Limitations and Scope}

This study is intentionally scoped as a \emph{controlled,
deployment-constrained case study} rather than a large-scale benchmark.
The protocol targets a real scanner pipeline where a lightweight model
must run in real time on consumer hardware; DeepLabV3-MobileNetV3
($\sim$11\,M parameters) represents the class of models deployed, not an
arbitrary legacy choice.
Our goal is not state-of-the-art segmentation, but rather understanding
how \emph{loss-function choice and evaluation methodology} interact
under extreme sparsity in this domain.
The narrow scope of this study facilitates an exhaustive, image-wise robustness analysis and the paired statistical tests presented above. Nevertheless, several limitations should be acknowledged.

\paragraph{Dataset size and domain.}
The data set comprises 34 original images (expanded to 374 by
augmentation) captured from a single whiteboard type and marker set.
This scale is representative of deployment-constrained scenarios where
labeled data collection is expensive; the three-seed protocol
mitigates over-fitting risk and the per-image analysis
(Table~\ref{tab:robustness}) provides fine-grained evidence.
The findings should be interpreted as a controlled study of the
loss-function design space under known conditions; cross-domain
generalization (different whiteboard brands, markers, ambient lighting)
is an empirical question for future work.

\paragraph{Single annotator.}
All masks were created by an annotator.  This ensures consistent
label semantics across the dataset, but inter-annotator agreement---particularly at thin-stroke boundaries where 1--2 pixel
discrepancies are common---remains unmeasured.

\paragraph{Architecture scope.}
Only DeepLabV3-MobileNetV3-Large was evaluated.  This choice was
deliberate: a lightweight fixed architecture isolates the effect of
the loss function from confounding architectural differences and
matches the deployment target (a real-time scanner
pipeline~on consumer GPUs).
Whether the observed loss-function ranking transfers to larger backbones
(e.g., ResNet-101, HRNet) or modern architectures such as
SegFormer~\cite{xie2021segformer} and
Mask2Former~\cite{cheng2022mask2former} is an open question.

\paragraph{Resolution ceiling.}
Deep models were evaluated at a maximum of $1536\times1152$,
substantially below the native image resolution
($3712\times2784$ / $3968\times2232$).
Training at native resolution was computationally prohibitive for our multi-seed ablation experiments on a single 10\,GB GPU. Patch-based training at native resolution therefore constitutes a natural direction for subsequent work.
Importantly, the classical baseline was evaluated at \emph{native}
resolution, which favors it in region metrics.
We explicitly articulate this asymmetry and acknowledge that the baseline’s advantage can diminish, or even disappear completely, if deep models are trained at higher input resolutions.

\section{Conclusion}
\label{sec:conclusion}

We have presented a boundary-aware evaluation protocol for binary
segmentation of whiteboard strokes under extreme class imbalance
(foreground $\approx$ 1.8\% of pixels).  The protocol combines
multi-seed training with nonparametric significance tests, per-image
robustness profiling, and boundary-level metrics to provide a more
complete picture of loss-function behavior than aggregate scores alone.
Our key findings are as follows.

\begin{enumerate}[nosep,leftmargin=*]
\item \textbf{Loss function selection is the primary determinant of performance.}
    Overlap-based loss functions (Dice, Dice+Focal, Tversky) increase \fone{} by more than 20 percentage points relative to distribution-based losses (cross-entropy, Focal), a difference that is statistically significant ($p < 0.001$, Wilcoxon signed-rank test) and associated with large effect sizes (median $\Delta$\fone{} $>$ 0.22).
  \item \textbf{Robustness is as critical as accuracy.}
    Dice-family loss functions yield interquartile ranges (IQR) $\leq 0.072$ and worst-case \fone{} scores $\geq 0.542$, whereas cross-entropy exhibits an IQR of 0.157 and a worst-case score of 0.230. The strongest classical baseline (Sauvola) attains the highest mean \fone{} (0.787) but only a worst-case value of 0.452, highlighting a clear trade-off between consistency and peak accuracy.
  \item \textbf{Thin strokes are disproportionately impacted.}
    Cross-entropy and Focal losses exhibit a core-to-thin \fone{} discrepancy of approximately 0.10, whereas Dice-family losses reduce this gap by half to roughly 0.06, indicating a more balanced treatment of fine-grained structures.
  \item \textbf{Boundary-sensitive metrics are indispensable.}
    BF1 and B-IoU reveal boundary-level trade-offs among Dice-family variants that cannot be captured by region-based metrics alone.
  \item \textbf{Input resolution constitutes a critical performance bottleneck.}
    Doubling the input resolution yields an increase of 13 points in \fone{} and 19 points in BF1, with even larger relative improvements observed for thin stroke regions.
\end{enumerate}

\subsection{Artifacts and Reproducibility}
\label{sec:artifacts}

All experiments are fully reproducible.  Every training run uses one
of three fixed random seeds ($\{42, 123, 7\}$) with deterministic
PyTorch settings (\texttt{torch.backends.cudnn.deterministic=True})
and identical data splits; all results are reported as
mean~$\pm$~standard deviation across the three seeds.
The public
repository\footnote{%
  \url{https://github.com/TheRealManual/OneNote-Whitebord-Scanner-Training}}
contains: (i)~the 34-image data set with binary masks and the augmentation
pipeline, (ii)~training code with the above seed and determinism
configuration, and
(iii)~evaluation scripts that regenerate every table and figure from
stored \texttt{test\_set\_evaluation.json} files.
Pre-trained weights for all 21 models (15~loss $\times$ 3~seed +
6~resolution $\times$ 3~seed) in PyTorch and ONNX formats are
available upon request.
All experiments were conducted on a single NVIDIA RTX~3080 (10\,GB).

\paragraph{Future work.}
Several directions merit further investigation:
(i)~training at higher spatial resolutions (or employing patch-based training pipelines at the native acquisition resolution) to reduce the performance gap relative to classical approaches,
(ii)~systematic evaluation of contemporary segmentation architectures, such as
SegFormer~\cite{xie2021segformer} and
Mask2Former~\cite{cheng2022mask2former}, to disentangle the influence of the loss function from that of the capacity of the backbone model.
(iii)~conducting controlled ablation studies of \emph{boundary-optimized} loss functions -- including InverseForm~\cite{borse2021inverseform},
Boundary loss~\cite{kervadec2019boundary} 
and clDice~\cite{shit2021cldice} -- under a unified experimental protocol,
and (iv)~expanding the data set to encompass a wider range of whiteboard brands, marker types and illumination conditions, thus enabling a more rigorous evaluation of cross-domain generalization.

\bibliographystyle{ACM-Reference-Format}
\bibliography{refs}

\end{document}